\documentclass{article}
\pdfoutput=1
\usepackage{microtype}
\usepackage{subfigure}
\usepackage{booktabs} 
\usepackage{url}

\usepackage{kbordermatrix}
\usepackage{booktabs}       

\usepackage{nicefrac}       
\usepackage{amsfonts}
\usepackage{microtype}      
\usepackage{subfigure}
\usepackage[pdftex]{graphicx}
\usepackage{amsmath}
\usepackage{algorithm}
\usepackage{multirow}
\usepackage{balance}
\usepackage{bbm}
\usepackage{multicol}

\usepackage[accepted]{icml2018}

\icmltitlerunning{Adaptive Memory Networks}

\newcommand{\mycomment}[1]{}
\newcommand{\ignore}[1]{}

\newcommand{\toolname}{Adaptive Memory Network} 
\newcommand{\toolshort}{AMN} 
\newcommand{\toolnames}{Adaptive Memory Networks}

\begin{document}

\twocolumn[
\icmltitle{Adaptive Memory Networks}




\begin{icmlauthorlist}
\icmlauthor{Daniel Li}{to}
\icmlauthor{Asim Kadav}{goo}
\end{icmlauthorlist}

\icmlaffiliation{to}{University of California-Berkeley, Berkeley, USA, Work done as a NEC Labs Intern.}
\icmlaffiliation{goo}{NEC Laboratories America, Princeton, USA}

\icmlcorrespondingauthor{Daniel Li}{li.daniel@berkeley.edu}
\icmlcorrespondingauthor{Asim Kadav}{asim@nec-labs.com}

\icmlkeywords{Machine Learning, ICML}

\vskip 0.3in
]



\printAffiliationsAndNotice{} 

\begin{abstract}

We present Adaptive Memory Networks (AMN) that processes input-question pairs to dynamically construct a network architecture optimized for lower inference times for Question Answering (QA) tasks. AMN processes the input story to extract entities and stores them in {\em memory banks}. Starting from a single bank, as the number of input entities increases, \toolshort\ learns to create new banks as the entropy in a single bank becomes too high. Hence, after processing an input-question(s) pair, the resulting network represents a hierarchical structure where entities are stored in different banks, distanced by question relevance.
At inference, one or few banks are used, creating a tradeoff between accuracy and performance.


\toolshort\ is enabled by dynamic networks that allow input dependent network creation and efficiency in dynamic mini-batching as well as our novel bank controller that allows learning discrete decision making with high accuracy. In our results, we demonstrate that \toolshort\ learns to create variable depth networks depending on task complexity and reduces inference times for QA tasks.


\end{abstract}

\section{Introduction}

Question Answering (QA) tasks are gaining significance due to their widespread applicability to recent commercial applications such as chatbots, voice assistants and even medical diagnosis~\cite{goodwin2016medical}. Furthermore, many existing natural language tasks can also be re-phrased as QA tasks. Providing faster inference times for QA tasks is crucial. Consumer device based question-answer services have hard timeouts for answering questions. For example, Amazon Alexa, a popular QA voice assistant, allows developers to extend the QA capabilities by adding new ``Skills'' as remote services~\cite{amazonalexa}. However, these service APIs are wrapped around hard-timeouts of 8 seconds which includes the time to transliterate the question to text on Amazon's servers and the round-trip transfer time of question and the answer from the remote service, and sending the response back to the device. Furthermore, developers are encouraged to provide a list of questions (``utterances'') apriori at each processing step to assist QA processing~\cite{amazonalexa}. We propose that apart from helping transliteration these questions can also provide hints for reducing inference times for QA tasks based on large knowledge bases.

Modeling QA tasks with LSTMs can be computationally expensive which is undesirable during inference. Memory networks, a class of deep networks with explicit addressable memory, have recently been used to achieve state of the art results on many QA tasks. Unlike LSTMs, where the number of parameters grows exponentially with the size of memory, memory networks are comparably parameter efficient and can learn over longer input sequences. However, they often require accessing all intermediate memory to answer a question. Furthermore, using focus of attention over the intermediate state using a list of questions does not address this problem. Soft attention based models compute a softmax over all states and hard attention models are not differentiable and can be difficult to train over a large state space. Previous work on improving inference over memory networks has focused on using unsupervised clustering methods to reduce the search space~\cite{chandar2016hierarchical, rae2016scaling}. Here, the memory importance is not {\em learned} and the performance of nearest-neighbor style algorithms is often comparable to a softmax operation over memories. 
%

To provide faster inference for long sequence-based inputs, we present \toolnames\ (\toolshort ), that constructs a memory network on-the-fly based on the input. Like past approaches to addressing external memory, \toolshort\ constructs the memory entity nodes dynamically. However, distinct from past approaches, \toolshort\ stores the entities from the input story in variable number of {\em memory banks}. The entities represent the hidden state of each word in the story while a memory bank is a collection of entities that are similar w.r.t the question. As the number of entities grow, and the entropy within a single bank becomes too high, our network learns to construct new memory banks and copies entities that are more relevant towards a single bank. Hence, by limiting the decoding step to a dynamic number of relevant memory banks, \toolshort\ achieves lower inference times. \toolshort\ is an end-to-end trained model with dynamic learned parameters for memory bank creation and movement of entities.

\begin{figure*}
\centering
\includegraphics[keepaspectratio=true,width=5in]{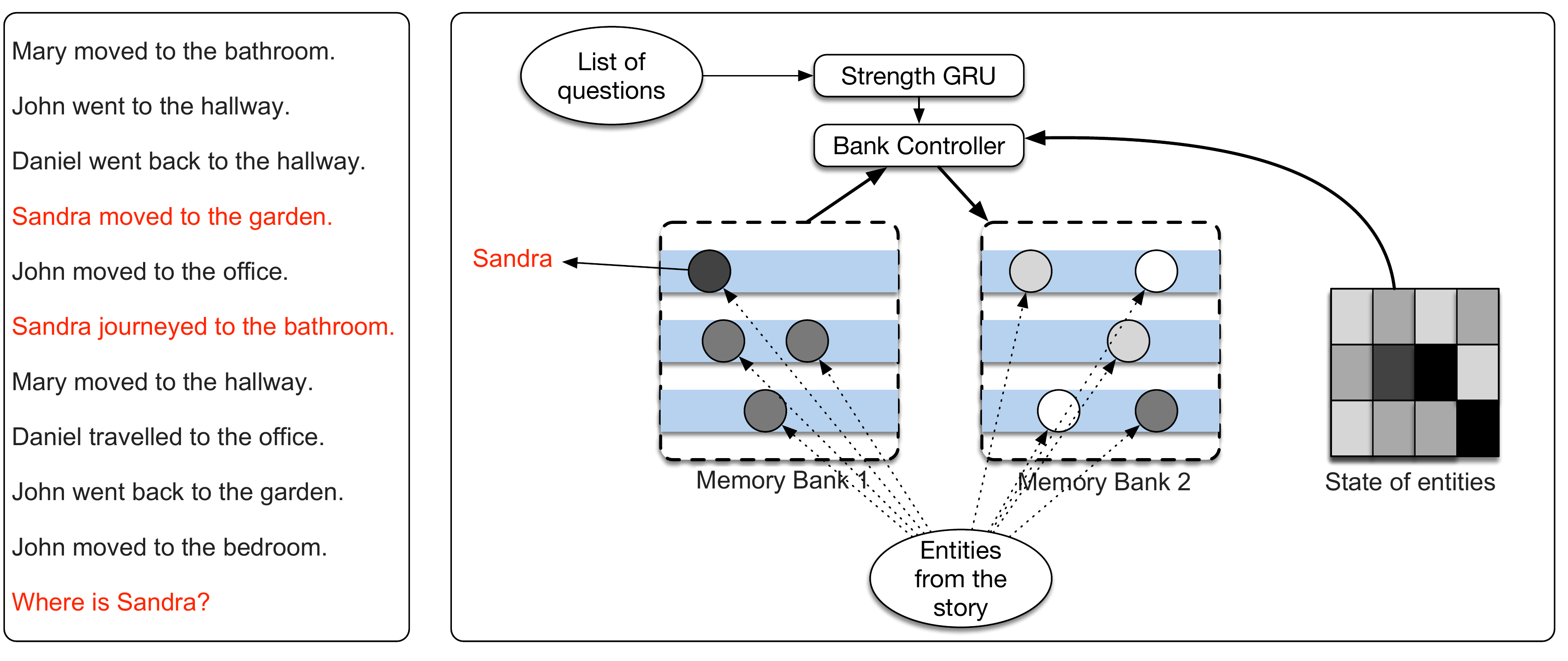}
\caption{\small \bf Overview of Adaptive memory networks. Multiple memory banks are created based on the story and input entities are moved  in them based on their relevance to the question. Inference is performed on a single (or less than all) banks most relevant to the question(s).}
\label{fig:overview}
\vspace{-0.2in}
\end{figure*}

Figure~\ref{fig:overview} demonstrates a simple QA task where \toolshort\ constructs two memory banks based on the input. During inference only the entities in the left bank are considered reducing inference times. To realize its goals, \toolshort\ introduces a novel bank controller that uses reparameterization trick to make discrete decisions with high accuracy while maintaining differentiability. Finally, \toolshort\ also models sentence structures on-the-fly and propagates update information for all entities that allows it to solve all $20$ bAbI tasks.


\section{Related Work}

\paragraph{Memory Networks:}Memory networks store the entire input sequence in memory and perform a softmax over hidden states to update the controller~\cite{weston2014memory, sukhbaatar15}. DMN+ connects memory to input tokens and updates them sequentially~\cite{xiong2016dynamic}. For inputs that consist of large number of tokens or entities, these methods can be expensive during inference. AMN stores entities with tied weights in different memory banks. By controlling the number of memory banks, AMN achieves low inference times with reasonable accuracy. Nearest neighbor methods have also been explored over memory networks. For example, Hierarchical Memory Networks separates the input memory into groups using the MIPS algorithm~\cite{chandar2016hierarchical}. However, using MIPS is as slow as a softmax operation, so the authors propose using an approximate MIPS that gives inferior performance. In contrast, \toolshort\  is end to end differentiable, and reasons which entities are important and constructs a network with dynamic depth.

Neural Turing Machine (NTM) consists of a memory bank and a differentiable controller that learns to read and write to specific locations~\cite{graves2014neural}. In contrast to NTMs, \toolshort\ memory bank controller is more coarse grained and the network learns to store entities in memory banks instead of specific locations. \toolshort\ uses a discrete bank controller that gives improved performance for bank controller actions over NTM's mechanisms. However, our design is consistent with the modeling studies of working memory by~\cite{hazy2006banishing} where the brain performs robust memory maintenance and may maintain multiple working representations for individual working tasks. Sparse access memory uses approximate nearest neighbors (ANN) to reduce memory usage in NTMs~\cite{rae2016scaling}. However, ANNs are not differentiable. AMN, uses a input specific memory organization that does not create sparse structures. This limits access during inference to specific entities reducing inference times.

Graph-based networks, (GG-NNs~\cite{li2015gated} and GGT-NNs~\cite{johnson2016learning}) use nodes with tied weights that are updated based on gated-graph state updates with shared weights over edges. However, unlike AMN, they require strong supervision over the input and teacher forcing to learn the graph structure. Furthermore, the cost of building and training these models is expensive and if every edge is considered at every time-step the amount of computation grows at the order of $O(N^3)$ where $N$ represents the number of nodes/entities. AMN does not use strong supervision but can solve tasks that require transitive logic by modeling sentence walks on the fly. EntNet constructs dynamic networks based on entities with tied weights for each entity~\cite{henaff2016tracking}. A key-value update system allows it to update relevant (learned) entities. However, Entnet uses soft-attention during inference to attend to all entities that incur high inference costs. To summarize, majority of the past work on memory networks uses softmax over memory nodes, where each node may represent input or an entity. In contrast, AMN learns to organize memory into various memory banks and performs decode over fewer entities reducing inference times.

\paragraph{Conditional Computation \& Efficient Inference:}\toolshort\ is also related to the work on conditional computation which allows part of networks to be active during inference improving computational efficiency ~\cite{bengio2015conditional}. Recently, this has been often accomplished using a gated mixture of experts ~\cite{eigen2013learning, shazeer2017outrageously}. AMN conditionally attends to entities in initial banks during inference improving performance. For faster inference using CNNs, pruning~\cite{obd, deepcompression}, low rank approximations~\cite{denton2014exploiting}, quantization and binarization~\cite{rastegari2016xnor} and other tricks to improve GEMM performance~\cite{vanhoucke2011improving} have been explored. For sequence based inputs, pruning and compression has been explored~\cite{giles1994pruning, see2016compression}. However, compression results in irregular sparsity that reduces memory costs but may not reduce computation costs. Adaptive computation time~\cite{graves2016adaptive} learns the number of steps required for inferring the output and this can also be used to reduce inference times~\cite{figurnov2016spatially}. \toolshort\ uses memory networks with dynamic number of banks to reduce computation costs. 

\paragraph{Dynamic networks:}Dynamic neural networks that change structure during inference have recently been possible due to newer frameworks such as Dynet and PyTorch. Existing work on pruning can be implemented using these frameworks to reduce inference times dynamically like dynamic deep network  demonstrates~\cite{liu2017dynamic}. AMN utilizes the dynamic architecture abilities to construct an input dependent memory network of variable memory bank depth and the dynamic batching feature to process a variable number of entities. Furthermore, unlike past work that requires an apriori number of fixed memory slots, \toolshort\ constructs them on-the-fly based on the input. The learnable discrete decision-making process can be extended to other dynamic networks which often rely on REINFORCE to make such decisions~\cite{liu2017dynamic}.

\paragraph{Neuroscience:}Our network construction is inspired by work on working memory representations. There is sufficient evidence for multiple, working memory representations in the human brain~\cite{hazy2006banishing}. Semantic memory~\cite{tulving1972episodic}, describes a hierarchical organization starting with relevant facts at the lowest level and progressively more complex and distant concepts at higher levels. \toolshort\ constructs entities from the input stories and stores the most relevant entities based on the question in the lowest level memory bank. Progressively higher level memory banks represent distant concepts (and not necessarily higher level concepts for \toolshort). Other work demonstrates organization of human memory in terms of ``priority structure'' where attention is a gate-keeper of working memory-guided by executive control's goals, plans, and intentions as in~\cite{watzl2017structuring}, similar in spirit to \toolshort's question guided network construction.



\section{Differentiable adaptive memory module}
\label{diffmem}



In this section, we describe the design process and motivation of our memory module. Our memory network architecture is created during inference time for every story. The architecture consists of different memory banks and each memory bank stores entities from the input story. Hence, a {\em memory entity} represents the hidden state of each entity (each word in our case) from the input story while a {\em memory bank} is a collection of entities. Intuitively, each memory bank stores entities that have a similar distance score from the question. 

At a high level, entities are gradually and recurrently copied through memory banks to \textit{filter} out irrelevant nodes such that in the final inference stage, fewer entities are considered by the decoder. Note that the word \textit{filter} implies a discrete decision and that recurrence implies time. If we were to perform a strict cut off and remove entities that appear to be irrelevant at each time step, learning the reasoning logic that requires previous entities that were cut off would not be possible. Thus, smoothed discretization is required.

We design filtering to be a two-stage pseudo-continuous process to simulate discrete cut offs ($\Pi_{move}, \Pi_{new}$), while keeping reference history. The overall memory ($M$) consists of multiple memory banks. A memory bank is a collection or group of entities ($m_{0...l}$), where $m_0$ denotes the initial and most general bank and $m_l$ denotes the most relevant bank. Note that $|l|$ is input dependent and learned. First, entities are moved from $m_0$ gradually towards $m_l$ based off of their individual relevance to the question and second, if $m_l$ becomes too saturated, $m_{l+1}$ is created. Operations in the external memory allowing for such dynamic restructuring and entity updates are described below. Note that these operations still maintain end to end differentiability.

\begin{enumerate}
    \item Memory bank creation ($\Pi_{new}$), which creates a new memory bank depending on the current states of entities $m_i$. If the entropy, or information contained (explained below), of $m_i$ is too high, $\Pi_{new}(m_i)$ will learn to create a new memory bank $m_{i+1}$ to reduce entropy.
    \item Moving entities across banks ($\Pi_{move}$), which determines which entities are relevant to the current question and move such entities to further (higher importance) memory banks.
    \item Adding/Updating entities in a bank ($\Pi_{au}$), which adds entities that are not yet encountered to the first memory bank $m_0$ or if the entity is already in $m_0$, the operation updates the entity state.
    \item Propagating changes across entities ($\Pi_{prop}$), which updates the entity states in memory banks based on node current states $\Pi_{prop}(M)$ and their semantic relationships. This is to communicate transitive logic. 
\end{enumerate}

Both $\Pi_{new}, \Pi_{move}$ require a discrete decision (refer to section 4.2.1.), and in particular, for $\Pi_{new}$ we introduce the notion of \textit{entropy}. That is to say if $m_i$ contains too many nodes (the entropy becomes too high), the memory module will learn to create a new bank $m_{i+1}$ and move nodes to $m_{i+1}$ to reduce entropy. By creating more memory banks, the model spreads out the concentration of information which in turn better discretizes nodes according to relevance.


\section{Adaptive Memory Networks}

\begin{figure*}
\centering
\includegraphics[keepaspectratio=true,width=5in]{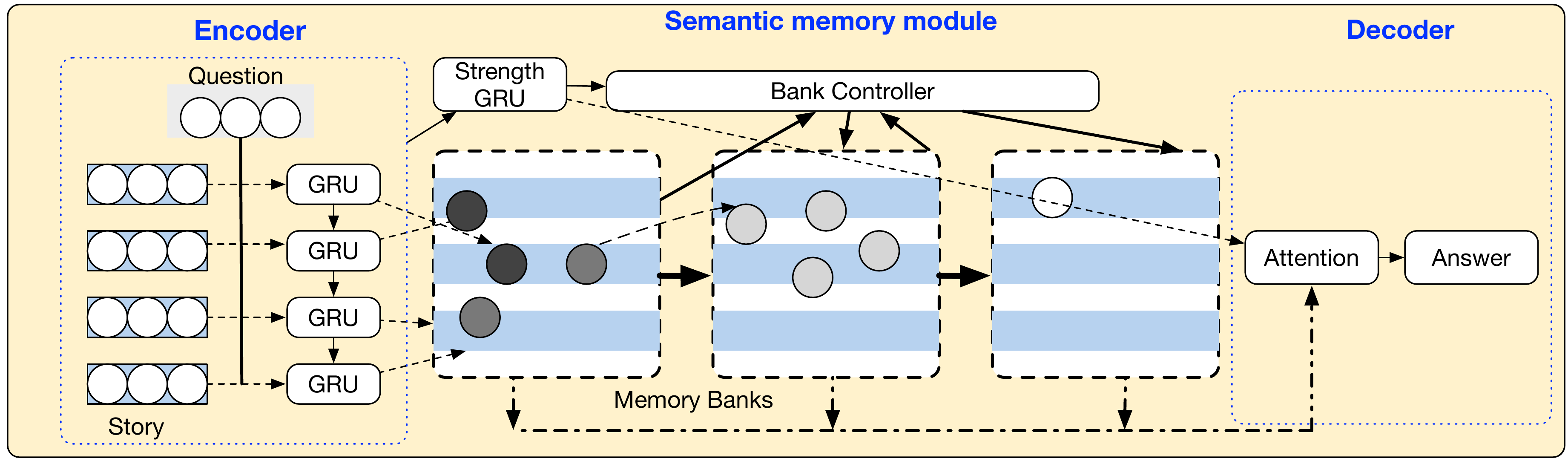}

\caption{\small \bf Adaptive memory networks.}

\label{fig:amn}
\vspace{-0.2in}
\end{figure*}

A high-level overview is shown in Figure~\ref{fig:amn}, followed by a mathematical detail of the model's modules. Our model adopts the encoder-decoder framework with an augmented adaptive memory module. For an overview of the algorithm, refer to Section \ref{subsec:algo}.

\paragraph{Notation and Problem Statement:}Given a story represented by $N$ {\bf input} sentences (or statements), i.e., $(l_1, \cdots, l_N)$, and a {\bf question} $q$, our goal is to generate an {\bf answer} $a$. Each sentence $l$ is a sequence of $N$ words, denoted as $(w_1, \cdots, w_{N_s})$, and a question is a sequence of $N_q$ words denoted as $(w_1, \cdots, w_{N_q})$. Throughout the model we refer to entities; these can be interpreted as a $3$-tuple of $\textbf{e}_{w} =$ (word ID $wi$, hidden state $\textbf{w}$, question relevance strength $\textbf{s}$). Scalars, vectors, matrices, and dot products are denoted by lower-case letters, boldface lower-case letters and boldface capital letters, and angled brackets respectively.

\subsection{Encoder}
The input to the model, starting with the encoder, are story-question input pairs. On a macro level, sentences $l_{1...N}$ are processed. On a micro level, words $w_{1...N_s}$ are processed within sentences.

For each $w_i \in l_i$, the encoder maps $w_i$ to a hidden representation and a question relevance strength $\in [0, 1]$. The word ID of $w_i$ is passed through a standard embedding layer and then encoded through an accumulation GRU. The accumulation GRU captures the entity states through time by adding the output of each GRU time step to its respective word, stored in a lookup matrix. The initial states of $\textbf{e}_w$ are set to this GRU output. Meanwhile, the question is also embedded and encoded in the same manner sans accumulation.

In the following, the subscripts $i,j$ are used to iterate through the total number of words in a statement and question respectively, $\textbf{D}$ stores the accumulation GRU output, and  $\textbf{w}_i$ is a GRU encoding output. The last output of the GRU will be referred to as $\textbf{w}_N, \textbf{w}_{N_q}$ for statements and questions.
\vspace{-0.2in}
\begin{center}
\small
\begin{align}
     \textbf{u}_{i}, \textbf{u}_{j} &= EMBED(wi_{i}), EMBED(wi_{j})\\
     \textbf{w}_{i} &= GRU(\textbf{u}_{i}, \textbf{w}_{i-1}) \\
   \textbf{D}[i] &\mathrel{{+}{=}} \textbf{w}_i\\
    \textbf{w}_{j} &= GRU(\textbf{u}_j, \textbf{w}_{j-1})
\end{align}

\end{center}
To compute the question relevance strength  $s \in [0, 1]$ for each word, the model uses GRU-like equations. The node strengths are first initialized to Xavier normal and the inputs are the current word states $\textbf{w}^{in}$, the question state $\textbf{w}_{N_q}$, and when applicable, the previous strength. Sentences are processed each time step $t$. 
\vspace{-0.2in}
\begin{center}
\small
\begin{align}
     \mathbf{z}^t &= \sigma(\mathbf{U}_z\mathbf{w}^{in} +\mathbf{W}_z\mathbf{w}_{N_q} + \mathbf{X}_z\mathbf{s}^{t-1})\\
     \mathbf{r}^t &= 1 - \sigma(\mathbf{U}_r\langle\mathbf{s}^{t-1}, \mathbf{w}_{N_q}\rangle) \\
    \widetilde{\mathbf{s}}^t &= \sigma(\mathbf{W}_h\mathbf{w}^{in} + \mathbf{U}_h(\mathbf{r}^t \odot \mathbf{s}^{t-1}))\\
    \mathbf{s}^t &= \mathbf{z}^t \odot \mathbf{s}^{t-1} + (1- \mathbf{z}^t)\odot  \widetilde{\mathbf{s}}^t
\end{align}
\end{center}

In particular, equation (6) shows where the model learns to lower the strengths of nodes that are not related the question. First, a dot product between the current word states and question state are computed for similarity (high correlation), then it is subtracted from a $1$ to obtain the dissimilarity. We refer to these operations as SGRU (Strength GRU) in Algorithm \ref{alg1}.


\subsection{Adaptive Memory Module}

The adaptive memory module recurrently restructures entities in a question relevant manner so the decoder can then consider fewer entities (namely, the question relevant entities) to generate an answer. The following operations are performed once per sentence.

\subsubsection{Memory Bank Controller}
As mentioned earlier, discrete decisions are difficult for neural networks to learn so we designed a specific memory bank controller $\Pi_{ctrl}$ for binary decision making. The model takes ideas from the reparameterization trick and uses custom backpropagation to maintain differentiability.


In particular, the adaptive memory module needs to make two discrete decisions on a $\{0, 1\}$ basis, one in $\Pi_{new}$ to create a new memory bank and the other in $\Pi_{move}$ to move nodes to a different memory bank. The model uses a scalar $p \in \{0, 1\}$ to parameterize a Bernoulli distribution where the realization $\textbf{H}$, is the decision the model makes. However, backpropagation through a random node is intractable, so the model detaches $\textbf{H}$ from the computation graph and introduces $\textbf{H}$ as a new node. Finally, $\textbf{H}$ is used as a mask to zero out entities in the discrete decision. Meanwhile, $p$ is kept in the computation graph and has a special computed loss (Section 4.4). The operations below will be denoted as $\Pi_{ctrl}$ and has two instances: one for memory bank creation $\Pi_{new}$ and one for moving entities across banks $\Pi_{move}$. In equation 9, depending on what $\Pi_{ctrl}$ is used for, $q$ is a polymorphic function and will take on a different operation and $*$ will be a different input. Examples of such are given in their respective sections (4.2.2.1, 4.2.2.2).






\begin{center}
\small
\begin{minipage}{.48\linewidth}
\begin{align}
 p &= q(*)
\end{align}
\end{minipage}
\hfill
\begin{minipage}{.5\linewidth}
\begin{align}
 \textbf{H} &= \text{Bernoulli}(p)
 \end{align}
\end{minipage}
\end{center}
\subsubsection{Memory Bank Operations}

\begin{enumerate}
\item {\bf Memory bank creation $\Pi_{new}$:} 
To determine when a new memory bank is created, in other words, if the current memory bank becomes too saturated, the memory bank controller (4.2.1.) will make a discrete decision to create a new memory bank. Here, $q$ (eq 9) is a fully connected layer and the input is the concatenation of all the current memory bank $m_i$'s entity states $[\textbf{w}_0 ... \textbf{w}_{i}] \in R^{1, n|\textbf{e}_w|}$. Intuitively, $q$ will learn a continuous decision that is later discretized by eq 10 based on entity states and the number of entities. Note this is only performed for the last memory bank.
\begin{center}
\begin{align}
  \Pi_{new}([\textbf{w}_0 ... \textbf{w}_{i}]) =
  \begin{cases}
  \textbf{M}.\text{new}() & \text{if $\mathbbm{1}(\Pi_{ctrl}([\textbf{w}_0 ... \textbf{w}_{i}]))$ else} \\
  \text{pass}
  \end{cases}
 \end{align}
\end{center}



\item {\bf Moving entities through memory banks:}
Similar to $\Pi_{new}$, individual entities' relevance scores are passed into the bank controller to determine $\textbf{H}$ as the input. The relevance score is computed by multiplying an entity state by its respective relevance $\in R^{n, |\textbf{e}_w|}$. Here, $q$ has a slight modification and is the identity function. Note that this operation can only be performed if there is a memory bank to move nodes to, namely if $m_{i+1}$ exists. Additionally, each bank has a set property where it cannot contain duplicate nodes, but the same node can exist in two different memory banks.
\begin{center}
\begin{align}
\Pi_{move}(\textbf{s}_{i} * \textbf{w}_{i}) &= m.\text{add}(\mathbbm{1}(\Pi_{ctrl}(\textbf{s}_{i} * \textbf{w}_{i}))) \quad\forall i \in m
\end{align}
\end{center}

\item {\bf Adding/Updating entities in a bank:}
Recall that entities are initially set to the output of $\textbf{D}$. However, as additional sentences are processed, new entities and their hidden states are observed. In the case of a new entity $\textbf{e}_w$, the entity is added to the first memory bank $m_0$. If the entity already exists in $m_0$, then $\textbf{e}_w$'s corresponding hidden state is updated through a GRU. This procedure is done for all memory banks.
\begin{center}
\begin{align}
  \Pi_{au}([\textbf{w}_0 ... \textbf{w}_{i}]) =
  \begin{cases}
  m_0\text{.add}(\textbf{e}^i_{w}) & \text{if $\textbf{e}_{w}^i \not \in m_0$ else} \\
  \textbf{w}^{t+1}_{i} = GRU(\textbf{w}_{N}, \textbf{w}_{i}^{t}) &  \forall m \in M
  \end{cases}
 \end{align}
\end{center}

\item {\bf Propagating updates to related entities:}
So far, entities exist as a bag of words model and the sentence structure is not maintained. This can make it difficult to solve tasks that require transitive reasoning over multiple entities. To track sentence structure information, we model semantic relationships as a directed graph stored in adjacency matrix $\textbf{A}$. As sentences are processed word by word, a directed graph is drawn progressively from $w_{0}...w_{i}...w_{N}$. If sentence $l_k$'s path contains nodes already in the current directed graph, $l_k$ will include said nodes in its path. After $l_k$ is added to $\textbf{A}$, the model propagates the new update hidden state information $\textbf{a}_i$ among all node states using a GRU. $\textbf{a}_i$ for each node $i$ is equal to the sum of the incoming edges' node hidden states. 

Additionally, we add a particular emphasis on $l_k$ to simulate recency. At face value, one propagation step of $\textbf{A}$ will only have a reachability of its immediate neighbor, so to reach all nodes, $\textbf{A}$ is raised to a consecutive power $r$ to reach and update each intermediate node. $r$ can be either the longest path in $\textbf{A}$ or a set parameter. Again, this is done within a memory bank for all memory banks. For entities that have migrated to another bank, the update for these entities is a no-op but propagation information as per the sentence structure is maintained. A single iteration is shown below:
\vspace{-0.2in}
\begin{center}
\small
\begin{minipage}{.48\linewidth}
\begin{align}
       \textbf{a} &= (\textbf{A}^r)^T[\textbf{w}_0...\textbf{w}_i]
\end{align}
\end{minipage}
\hfill
\begin{minipage}{.5\linewidth}
\begin{align}
 \textbf{w}^{t} = GRU(\textbf{a}, \textbf{w}^{t-1})
\end{align}
\end{minipage}
\end{center}
When nodes are transferred across banks, $\textbf{A}$ is still preserved. If intermediate nodes are removed from a path, a transitive closure is drawn if possible. 


\end{enumerate}
After these steps are finished at the end of a sentence, namely, the memory unit has reasoned through how large (number of memory banks) the memory should be and which entities are relevant at the current point in the story, all entities are passed through the strength modified GRU (4.1, eq 5-8) to recompute their question relevance (relevance score).

\subsection{Decode}
After all sentences $l_{1...N}$ are ingested, the decode portion of the network learns to interpret the results from the memory banks. The network iterates through the memory banks using a standard attention mechanism. To force the network to understand the question importance weighting, the model uses an exponential function $d$ to weight important memory banks higher. $\textbf{C}_m$ are the hidden states contained in memory $m$, $\textbf{s}_{m}$ are the relevance strengths of memory bank $m$, $\textbf{w}_{N_q}$ is the question hidden state, $\textbf{ps}$ is the attention score, $r,h$ are learned weight masks, $\textbf{g}$ are the accumulated states, and $\textbf{l}$ is the final logits prediction. During \textit{inference}, fewer memory banks are considered.
\begin{center}
\small
\begin{align}
 \textbf{C}_m &= \textbf{s}_m \cdot [\textbf{w}_0, ... \textbf{w}_i]\quad \forall i \in m \\
 \textbf{ps} &= Softmax(\langle \textbf{C}_m, \textbf{w}_{N_q} \rangle) \\ 
\textbf{g} &\mathrel{{+}{=}} d(\textstyle\langle \textbf{C}_m, \textbf{ps} \rangle)\\
\hat{\textbf{L}} &= r(
\text{PReLU}(h(\textbf{g}) + \textbf{w}_{N_q}) \quad \text{if $m$ is last}
\end{align}
\end{center}

\subsection{Loss}
Loss is comprised of two parts, answer loss, which is computed from the given annotations, and secondary loss (from $\Pi_{new}$, $\Pi_{move}$), which is computed from sentence and story features at each sentence time step $l_{0...N}$.
Answer loss is standard cross entropy at the end of the story after $l_N$ is processed.
$$\mathcal{L}_p(\hat{\textbf{L}}) = \text{CrossEntropy}(\hat{\textbf{L}}, \textbf{L})$$
After each sentence $l_i$, the node relevance $\textbf{s}_{l_i}$ is enforced by computing the expected relevance $\mathbb{E}[\textbf{s}_{l_i}]$. $\mathbb{E}[\textbf{s}]$ is determined by nodes that are connected to the answer node $\textbf{a}$ in a directed graph; words that are connected to $\textbf{a}$ are relevant to $\textbf{a}$. They are then weighted with a deterministic function of distance from $\textbf{a}$.
$$\mathcal{L}_s(\textbf{s}) = D_{KL}(\textbf{s}_{l_i}||\mathbb{E}[\textbf{s}_{l_i}])$$
Additionally, bank creation is kept in check by constraining $p_{l_i}$ w.r.t. the expected number of memory banks. The expected number of memory banks can be thought of as a geometric distribution $\sim \text{Geometric}(\hat{p}_{l_i})$ parameterized by $\hat{p}_{l_i}$, a hyperparameter. Typically, at each sentence step $\hat{p}$ is raised to the inverse power of the current sentence step to reflect the amount of information ingested. Intuitively, this loss ensures banks are created when a memory bank contains too many nodes. On the other hand, the learned mask $q$ (eq. 9) enables the model to weight certain nodes a higher entropy to prompt bank creation. Through these two dependencies, the model is able to simulate bank creation as a function of the number of nodes and the type of nodes in a given memory bank.
$$\mathcal{L}_{b}(p_{l_i}) = D_{KL}(p_{l_i}||\hat{p}^{\frac{1}{\beta|l_i|}})$$
All components combined, the final loss is given in the following equation
$$\mathcal{L}_{total} = \mathcal{L}_{p}(\hat{\textbf{L}}) + \sum_{i = 1}^{|l_n|}(\mathcal{L}^i_s(\textbf{s}) + \mathcal{L}_{b}^i(p))$$


\section{Evaluation}

In this section, we evaluate \toolshort\ accuracy and inference times on the bAbI dataset~\cite{weston2015towards} and extended bAbI tasks dataset. We compare our performance with Entnet~\cite{henaff2016tracking}, which recently achieved state of the art results on the bAbi dataset. For accuracy measurements, we also compare with DMN+ and encoder-decoder methods. Finally we discuss the time trade offs between \toolshort\ and current SOTA methods. The portion regarding inference times are not inclusive of story ingestion. We summarize our experiments results as follows:

\begin{itemize}

\item We are able to solve all bAbi tasks using \toolshort. Furthermore, \toolshort\ is able to reason important entities and propagate them to the final memory bank allowing for 48\% fewer entities examined during inference.

\item We construct extended bAbI tasks to evaluate \toolshort\ behavior. First, we extend Task 1 for \textit{multiple questions} in order to gauge performance in a more robust manner. For example, if a reasonable set of questions are asked (where reasonable means that collectively they do not require all entities to answer implying entities can be filtered out), will the model still sufficiently reason through entities. We find that our network is able to reason useful entities for both tasks and store them in the final memory bank. Furthermore, we also scale bAbI for a large number of entities and find that \toolshort\ provides additional benefits at scale since only relevant entities are stored in the final memory bank.

\end{itemize}

\subsection{Experiment settings}

 We implement our network in PyTorch~\cite{paszke2017pytorch}. We initialize our model using Xavier initialization, and the word embeddings utilize random uniform initialization ranging from $-\sqrt{3}$ to $\sqrt{3}$. The learning rate is set as $0.001$ initially and updated with a learning rate scheduler. $\mathbb{E}[\textbf{s}]$ contains nodes in the connected components of $\textbf{A}$ containing the answer node $\textbf{a}$ which has relevance scores sampled from a Gaussian distribution centered at $0.75$ with a variance of $0.05$ (capped at 1). Nodes that are not in the connected component containing $\textbf{a}$ are similarly sampled from a Gaussian centered from $0.3$ with a variance of $0.1$ (capped at 0). $\hat{p}_{l_i}$ is initially set to $0.8$ and $\beta$ varies depending on the story length from $0.1 \leq \beta \leq 0.25$. Note that for transitive tasks, $\hat{p}_{l_i}$ is set to 0.2.  We train our models using the Adam optimizer~\citep{kingma2014adam}.



\subsection{bAbi dataset}
The bAbI task suite consists of 20 reasoning tasks that include deduction, induction, path finding etc. Results are from the following parameters: $\leq200$ epochs, best of 10 runs. Table~\ref{table:babiaccuracy} shows the accuracy and Table ~\ref{table:inference} shows the inference performance in terms of the number of entities examined. A task is considered passed if the error rate is less than 5\%. 

We find that \toolshort\ creates $1 - 6$ memory banks for different tasks. We also find that $8$ tasks can be solved by looking at just one memory bank and $14$ tasks can be solved with half the total number of memory banks. Lastly, all tasks can be solved by examining less than or equal the total number of entities $(e \in M \leq |V| + \epsilon)$\footnote{The entities used to construct an answer \textit{and pass the task} are examined as the sum of all entities across the $M$ which is usually $O(|V|)$. However, this is within an error margin of $6\%$ more entities on some experiments, and thus we included an $\epsilon$ term.}. Tasks that cannot be solved in fewer than half the memory banks either require additional entities due to transitive logic or have multiple questions. For transitive logic, additional banks could be required as an relevant nodes may be in a further bank. However, this still avoids scanning all banks. In the case of multiple questions, all nodes may become necessary to construct all answers. We provide additional evaluation in Appendix to examine memory bank behavior for certain tasks.



\begin{table*} [ht]
	\small
	\begin{center}
	\vspace{-0.15cm}
		\begin{tabular}{l c c c c c}
			\hline
			Task & AMN & Entnet & DMN+ & MemN2N & EncDec\\
			\hline
			1 - Single Supporting Fact & 
			\textbf{0.0} &  0.0 & 0.0 & 0.0 & 52.0\\
			2 - Two Supporting Facts & 
			\textbf{2.1} & 0.1 & 0.3 & 0.3 & 66.1  \\
			3 - Three Supporting Facts & 
			4.7 & 4.1 & 1.1 & 2.1 & 71.9 \\
			4 - Two Arg. Relations & 
		   \textbf{0.0} & 0.0 & 0.0 & 0.0 & 29.2     \\
			5 - Three Arg. Relations & 
			\textbf{2.7} & 0.3 & 0.5 & 0.8 & 14.3   \\
			6 - Yes/No Questions & 
			3.1 & 0.2 & 0.0 & 0.1 & 31.0       \\
			7 - Counting & 
			0.0 & 0.0 & 2.4 & 2.0 & 21.8 \\
			8 - Lists/Sets & 
			0.0 & 0.5 & 0.0 & 0.9 & 27.6 \\
			9 - Simple Negation & 
			1.3 & 0.1 & 0.0 & 0.3 & 36.4\\
			10 - Indefinite Knowledge &
		\textbf{1.2} & 0.6 & 0.0 & 0.0 & 36.4\\
			11 - Basic Coreference & 
			 \textbf{2.7} & 0.3 & 0.0 & 0.1 & 31.7 \\
			12 - Conjunction & 
			\textbf{2.2} & 0.0 & 0.0 &0.0 & 35.0 \\
			13 - Compound Coref. & 
			4.6 & 1.3 & 0.0 & 0.0 & 6.80 \\
			14 - Time Reasoning & 
			 \textbf{2.1} & 0.0 & 0.2 & 0.1 & 67.2\\
			15 - Basic Deduction & 
			\textbf{1.8} & 0.0 & 0.0 & 0.0 & 62.2 \\
			16 - Basic Induction & 
			4.2 & 0.0 & 45.3 & 51.8 & 54.0 \\
			17 - Positional Reasoning & 
			\textbf{4.3} & 0.5 & 4.2 & 18.6 & 43.1 \\
			18 - Size Reasoning & 2.0 & 0.3 & 2.1 & 5.3 & 6.60 \\
			19 - Path Finding & 2.4 & 2.3 & 0.0 & 2.3  & 89.6 \\
			20 - Agent’s Motivations &\textbf{ 0.0} & 0.0 & 0.0 & 0.0 & 2.30 \\
			\hline
			No. of failed tasks ($>$5\%)  & \textbf{0} & \textbf{0} & 5 & 6 & 20 \\
			\hline
		\end{tabular}
		\caption{Performance comparison of various models in terms of test error rate (\%) and the number of failed tasks on the bAbI dataset. The bold task scores are where AMN can solve the task using only $1$ memory bank. }
		\label{table:babiaccuracy}
		\vspace{-0.5cm}
	\end{center}
\end{table*}

\paragraph{Inference performance}

\begin{table*} [ht]
	\small
	\begin{center}
		\begin{tabular}{l c c c c c}
			\hline
			Task & Created Banks (Rounded Average) & Required Banks & Ratio ($\frac{e \in M}{|V|}$)\\
			\hline
			1 - Single Supporting Fact & 3 & 1 & 0.22\\
			2 - Two Supporting Facts & 5 & 1 & 0.41\\
			4 - Two Arg. Relations & 2 & 1 & 0.70\\
			7 - Counting & 5 & 2 & 0.81\\
			10 - Indefinite Knowledge & 1 & 1 & 1.00\\
			11 - Basic Coreference & 3 & 1 & 0.43\\
			12 - Conjunction & 2 & 1 & 0.37\\
			14 - Time Reasoning & 3 & 1 & 0.60\\
			15 - Basic Deduction & 1 & 1 & 1.00\\
			16 - Basic Induction & 2 & 2 & \textit{1.06}\\
			17 - Positional Reasoning & 1 & 1 & 1.00\\
			18 - Size Reasoning & 3 & 2 & 0.82\\
			19 - Path Finding & 2 & 2 & \textit{1.05}\\
			20 - Agent’s Motivations & 2 & 1 & 0.26\\	\hline
			Extended bAbi \\\hline
			1 - Single Supporting Fact, 100 Entities & 6 & 1 & .13 \\
			1 - Single Supporting, Multiple Questions & 3 & 1 & .38 \\ 
			\hline
		\end{tabular}
		\caption{Memory bank analysis of indicative tasks.}
		\label{table:inference}
		\vspace{-0.2in}
	\end{center}
\end{table*}

Table ~\ref{table:inference} shows the number of banks created and required to solve a task, as well as the ratio of entities examined to solve the task. Table 3 shows the complexity of AMN and other SOTA models. Entnet uses an empirically selected parameter, typically set to the number of vocabulary words. GGT-NN uses the number of vocabulary words and creates new $k$ new nodes intermittently per sentence step. 

For tasks where nodes are easily separable where nodes are clearly irrelevant to the question(s), AMN is able to successfully reduce the number of nodes examined. However for tasks that require more information, such as counting (Task 7), the model is still able to obtain the correct answer without using all entities. Lastly, transitive logic tasks where information is difficult to separate due to dependencies of entities, the model creates very few banks (1 or 2) and uses all nodes to correctly generate an answer. We note that in the instance where the model only creates one bank, it is very sparse, containing only one or two entities.

Because variations in computation times in text are minute, the number of entities required to construct an answer are of more interest as they directly correspond to the number of computations required. Additionally, due to various implementations of current models, their run times can significantly vary. However, for the comparison of inference times, AMN's decoder and EntNet's decoder are highly similar and contain roughly the same number of operations. We provide inference wall-clock time performance and memory bank behavior in supplementary behavior for representative tasks.




\subsection{Extended bAbi tasks}
We extend the bAbI tasks by adding additional entities and sentences and adding multiple questions for a single story, for Task 1. 

\paragraph{Scaled Task 1:}We increase the the number of entities to $100$ entities in the task generation system instead of existing $38$. We also extend the story length to $90$ to ensure new entities are referenced. We find that \toolshort\ creates $6$ memory banks and the ratio of entities in the final banks versus the overall entities drops to $0.13$ given the excess entities that are not referenced in the questions. 

\paragraph{Multiple questions:} We also augment the tasks with multiple questions to understand if \toolshort\ can handle when a story has multiple questions associated with it. We extend our model to handle multiple questions at once to limit re-generating the network for every question. To do so, we modify bAbi to generate several questions per story for tasks that do not currently have multiple questions. For single supporting fact (Task 1), the model creates $3$ banks and requires 1 bank to successfully pass the task. \textit{Furthermore}, the ratio of entities required to pass the task only increases by $0.16$ for a total of $0.38$.



\section{Conclusion and Future Work}

In this paper, we present \toolname\ that learns to adaptively organize the memory to answer questions with lower inference times. Unlike NTMs which learn to read and write at individual memory locations, \toolname\ demonstrates a novel design where the learned memory management is coarse-grained that is easier to train.Through our experiments, we demonstrate that \toolshort\ can learn to reason, construct, and sort memory banks based on relevance over the question set. 

\toolshort\ presents a new design paradigm in memory networks. Unlike NTMs, where the network learns where to read/write to fine-grained address information, \toolshort\ only learns to write input entities to coarse-grained banks. As a result, \toolshort\ is easier to train (e.g. does not require curriculum learning like NTMs) and does not require a separate sparsification mechanism like approximate nearest neighbors for inference efficiency. Apart from saving inference times, AMN can learn to reason which specific entities contribute towards the final answer improving interpretability.

\balance

\nocite{langley00}


\begin{thebibliography}{30}
\providecommand{\natexlab}[1]{#1}
\providecommand{\url}[1]{\texttt{#1}}
\expandafter\ifx\csname urlstyle\endcsname\relax
  \providecommand{\doi}[1]{doi: #1}\else
  \providecommand{\doi}{doi: \begingroup \urlstyle{rm}\Url}\fi

\bibitem[Amazon(2017)]{amazonalexa}
Amazon.
\newblock Alexa skills kit.
\newblock In \emph{\url{https://developer.amazon.com/alexa-skills-kit}}, 2017.

\bibitem[Bengio et~al.(2015)Bengio, Bacon, Pineau, and
  Precup]{bengio2015conditional}
Bengio, Emmanuel, Bacon, Pierre-Luc, Pineau, Joelle, and Precup, Doina.
\newblock Conditional computation in neural networks for faster models.
\newblock \emph{arXiv preprint arXiv:1511.06297}, 2015.

\bibitem[Chandar et~al.(2016)Chandar, Ahn, Larochelle, Vincent, Tesauro, and
  Bengio]{chandar2016hierarchical}
Chandar, Sarath, Ahn, Sungjin, Larochelle, Hugo, Vincent, Pascal, Tesauro,
  Gerald, and Bengio, Yoshua.
\newblock Hierarchical memory networks.
\newblock \emph{arXiv preprint arXiv:1605.07427}, 2016.

\bibitem[Denton et~al.(2014)Denton, Zaremba, Bruna, LeCun, and
  Fergus]{denton2014exploiting}
Denton, Emily~L, Zaremba, Wojciech, Bruna, Joan, LeCun, Yann, and Fergus, Rob.
\newblock Exploiting linear structure within convolutional networks for
  efficient evaluation.
\newblock In \emph{Advances in Neural Information Processing Systems}, pp.\
  1269--1277, 2014.

\bibitem[Eigen et~al.(2013)Eigen, Ranzato, and Sutskever]{eigen2013learning}
Eigen, David, Ranzato, Marc'Aurelio, and Sutskever, Ilya.
\newblock Learning factored representations in a deep mixture of experts.
\newblock \emph{arXiv preprint arXiv:1312.4314}, 2013.

\bibitem[Figurnov et~al.(2016)Figurnov, Collins, Zhu, Zhang, Huang, Vetrov, and
  Salakhutdinov]{figurnov2016spatially}
Figurnov, Michael, Collins, Maxwell~D, Zhu, Yukun, Zhang, Li, Huang, Jonathan,
  Vetrov, Dmitry, and Salakhutdinov, Ruslan.
\newblock Spatially adaptive computation time for residual networks.
\newblock \emph{arXiv preprint arXiv:1612.02297}, 2016.

\bibitem[Giles \& Omlin(1994)Giles and Omlin]{giles1994pruning}
Giles, C~Lee and Omlin, Christian~W.
\newblock Pruning recurrent neural networks for improved generalization
  performance.
\newblock \emph{IEEE transactions on neural networks}, 5\penalty0 (5):\penalty0
  848--851, 1994.

\bibitem[Goodwin \& Harabagiu(2016)Goodwin and Harabagiu]{goodwin2016medical}
Goodwin, Travis~R and Harabagiu, Sanda~M.
\newblock Medical question answering for clinical decision support.
\newblock In \emph{Proceedings of the 25th ACM International on Conference on
  Information and Knowledge Management}, pp.\  297--306. ACM, 2016.

\bibitem[Graves(2016)]{graves2016adaptive}
Graves, Alex.
\newblock Adaptive computation time for recurrent neural networks.
\newblock \emph{arXiv preprint arXiv:1603.08983}, 2016.

\bibitem[Graves et~al.(2014)Graves, Wayne, and Danihelka]{graves2014neural}
Graves, Alex, Wayne, Greg, and Danihelka, Ivo.
\newblock Neural turing machines.
\newblock \emph{arXiv preprint arXiv:1410.5401}, 2014.

\bibitem[Han et~al.(2016)Han, Mao, and Dally]{deepcompression}
Han, Song, Mao, Huizi, and Dally, William~J.
\newblock {Deep Compression: Compressing Deep Neural Networks with Pruning,
  Trained Quantization and Huffman Coding}.
\newblock In \emph{ICLR}, 2016.

\bibitem[Hazy et~al.(2006)Hazy, Frank, and O’Reilly]{hazy2006banishing}
Hazy, Thomas~E, Frank, Michael~J, and O’Reilly, Randall~C.
\newblock Banishing the homunculus: making working memory work.
\newblock \emph{Neuroscience}, 139\penalty0 (1):\penalty0 105--118, 2006.

\bibitem[Henaff et~al.(2017)Henaff, Weston, Szlam, Bordes, and
  LeCun]{henaff2016tracking}
Henaff, Mikael, Weston, Jason, Szlam, Arthur, Bordes, Antoine, and LeCun, Yann.
\newblock Tracking the world state with recurrent entity networks.
\newblock \emph{Proceedings of the International Conference on Learning
  Representations}, 2017.

\bibitem[Johnson(2017)]{johnson2016learning}
Johnson, Daniel~D.
\newblock Learning graphical state transitions.
\newblock In \emph{ICLR}, 2017.

\bibitem[Kingma \& Ba(2014)Kingma and Ba]{kingma2014adam}
Kingma, Diederik and Ba, Jimmy.
\newblock Adam: A method for stochastic optimization.
\newblock \emph{arXiv preprint arXiv:1412.6980}, 2014.

\bibitem[Le~Cun et~al.(1989)Le~Cun, Denker, and Solla]{obd}
Le~Cun, Yann, Denker, John~S, and Solla, Sara~A.
\newblock Optimal brain damage.
\newblock In \emph{NIPS}, 1989.

\bibitem[Li et~al.(2015)Li, Tarlow, Brockschmidt, and Zemel]{li2015gated}
Li, Yujia, Tarlow, Daniel, Brockschmidt, Marc, and Zemel, Richard.
\newblock Gated graph sequence neural networks.
\newblock \emph{arXiv preprint arXiv:1511.05493}, 2015.

\bibitem[Liu \& Deng(2017)Liu and Deng]{liu2017dynamic}
Liu, Lanlan and Deng, Jia.
\newblock Dynamic deep neural networks: Optimizing accuracy-efficiency
  trade-offs by selective execution.
\newblock \emph{arXiv preprint arXiv:1701.00299}, 2017.

\bibitem[Paszke et~al.(2017)Paszke, Gross, and Chintala]{paszke2017pytorch}
Paszke, Adam, Gross, Sam, and Chintala, Soumith.
\newblock Pytorch, 2017.

\bibitem[Rae et~al.(2016)Rae, Hunt, Danihelka, Harley, Senior, Wayne, Graves,
  and Lillicrap]{rae2016scaling}
Rae, Jack, Hunt, Jonathan~J, Danihelka, Ivo, Harley, Timothy, Senior, Andrew~W,
  Wayne, Gregory, Graves, Alex, and Lillicrap, Tim.
\newblock Scaling memory-augmented neural networks with sparse reads and
  writes.
\newblock In \emph{Advances in Neural Information Processing Systems}, pp.\
  3621--3629, 2016.

\bibitem[Rastegari et~al.(2016)Rastegari, Ordonez, Redmon, and
  Farhadi]{rastegari2016xnor}
Rastegari, Mohammad, Ordonez, Vicente, Redmon, Joseph, and Farhadi, Ali.
\newblock Xnor-net: Imagenet classification using binary convolutional neural
  networks.
\newblock In \emph{European Conference on Computer Vision}, pp.\  525--542.
  Springer, 2016.

\bibitem[See et~al.(2016)See, Luong, and Manning]{see2016compression}
See, Abigail, Luong, Minh-Thang, and Manning, Christopher~D.
\newblock Compression of neural machine translation models via pruning.
\newblock \emph{Proceedings of the SIGNLL Conference on Computational Natural
  Language Learning (CoNLL)}, 2016.

\bibitem[Shazeer et~al.(2017)Shazeer, Mirhoseini, Maziarz, Davis, Le, Hinton,
  and Dean]{shazeer2017outrageously}
Shazeer, Noam, Mirhoseini, Azalia, Maziarz, Krzysztof, Davis, Andy, Le, Quoc,
  Hinton, Geoffrey, and Dean, Jeff.
\newblock Outrageously large neural networks: The sparsely-gated
  mixture-of-experts layer.
\newblock \emph{arXiv preprint arXiv:1701.06538}, 2017.

\bibitem[Sukhbaatar et~al.(2015)Sukhbaatar, szlam, Weston, and
  Fergus]{sukhbaatar15}
Sukhbaatar, Sainbayar, szlam, arthur, Weston, Jason, and Fergus, Rob.
\newblock End-to-end memory networks.
\newblock pp.\  2440--2448, 2015.

\bibitem[Tulving et~al.(1972)]{tulving1972episodic}
Tulving, Endel et~al.
\newblock Episodic and semantic memory.
\newblock \emph{Organization of memory}, 1:\penalty0 381--403, 1972.

\bibitem[Vanhoucke et~al.(2011)Vanhoucke, Senior, and
  Mao]{vanhoucke2011improving}
Vanhoucke, Vincent, Senior, Andrew, and Mao, Mark~Z.
\newblock Improving the speed of neural networks on cpus.
\newblock In \emph{Proc. Deep Learning and Unsupervised Feature Learning NIPS
  Workshop}, volume~1, pp.\ ~4, 2011.

\bibitem[Watzl(2017)]{watzl2017structuring}
Watzl, Sebastian.
\newblock \emph{Structuring mind: The nature of attention and how it shapes
  consciousness}.
\newblock Oxford University Press, 2017.

\bibitem[Weston et~al.(2014)Weston, Chopra, and Bordes]{weston2014memory}
Weston, Jason, Chopra, Sumit, and Bordes, Antoine.
\newblock Memory networks.
\newblock \emph{arXiv preprint arXiv:1410.3916}, 2014.

\bibitem[Weston et~al.(2015)Weston, Bordes, Chopra, Rush, van Merri{\"e}nboer,
  Joulin, and Mikolov]{weston2015towards}
Weston, Jason, Bordes, Antoine, Chopra, Sumit, Rush, Alexander~M, van
  Merri{\"e}nboer, Bart, Joulin, Armand, and Mikolov, Tomas.
\newblock {Towards AI-complete question answering: A set of prerequisite toy
  tasks}.
\newblock \emph{arXiv preprint arXiv:1502.05698}, 2015.

\bibitem[Xiong et~al.(2016)Xiong, Merity, and Socher]{xiong2016dynamic}
Xiong, Caiming, Merity, Stephen, and Socher, Richard.
\newblock Dynamic memory networks for visual and textual question answering.
\newblock In \emph{International Conference on Machine Learning}, pp.\
  2397--2406, 2016.

\end{thebibliography}
\bibliographystyle{icml2018}
\newpage
\appendix
\nobalance
\section{Appendix}
\subsection{Decode overhead}
\begin{table}[t]
    \centering

    \small
    \begin{tabular}{cccc}
    Method     & Complexity 	\\ \hline
    Entnet (\cite{henaff2016tracking})        & $O(|V|)$ \\ 
    GGT-NN (\cite{johnson2016learning})   &  $O(|V| + kS)$	 \\ 
    \textbf{AMN (ours)} &  $O(\alpha|V|)) : 0 < \alpha < 1 + \epsilon$    \\\hline
    \end{tabular}
        \caption{
        Comparison of decode complexity for AMN, Entnet and GGT-NN.}
    \label{table:amn-eff}
\end{table}
\begin{table*} [ht]
    \small
    \begin{center}
        \begin{tabular}{l c c c c }
            \hline
            Task & Created Banks & Required Banks & Baseline (All banks) & AMN (Required banks) \\
            \hline
            1 - Single Supporting Fact & 3 & 1 & 2.15 s & 0.6 s\\
            2 - Two Supporting Facts & 5 & 1 & 15.8 s & 3.2 s\\
            7 - Counting & 5 & 2 & 21 s & 6.0 s\\
            \hline
        \end{tabular}
        \caption{Memory bank wall clock times for representative tasks for 1000 examples (time in secs).}
        \label{table:inference}
        \vspace{-0.1cm}
    \end{center}
\end{table*}

We compare the computations costs during the decode operation during inference for solving the extended bAbi task. We compute the overheads for \toolshort\, Entnet~\cite{henaff2016tracking} and GGT-NN. Table~\ref{table:amn-eff} gives the decode comparisons between \toolshort, Entnet and GGT-NN. Here, $|V|$ represents to the total number of entities for all networks. GGT-NN can dynamically create nodes and $k$ k is hyper parameter the new nodes created for $S$ sentences in input story. $\alpha$ is the percent of entities stored in the final bank w.r.t to the total entities for \toolshort.

We compare the wall clock execution times for three tasks within bAbI for 1000 examples/task. We compare the wall-clock times for three tasks. We compare the inference times of considering all banks (and entities) versus the just looking at the passing banks as required by AMN. We find that AMN requires fewer banks and as a consequence fewer entities and saves inference times.

\subsection {Propagation Example}
\label{prop}
\begin{figure}
\centering
\includegraphics[keepaspectratio=true,width=3.2in]{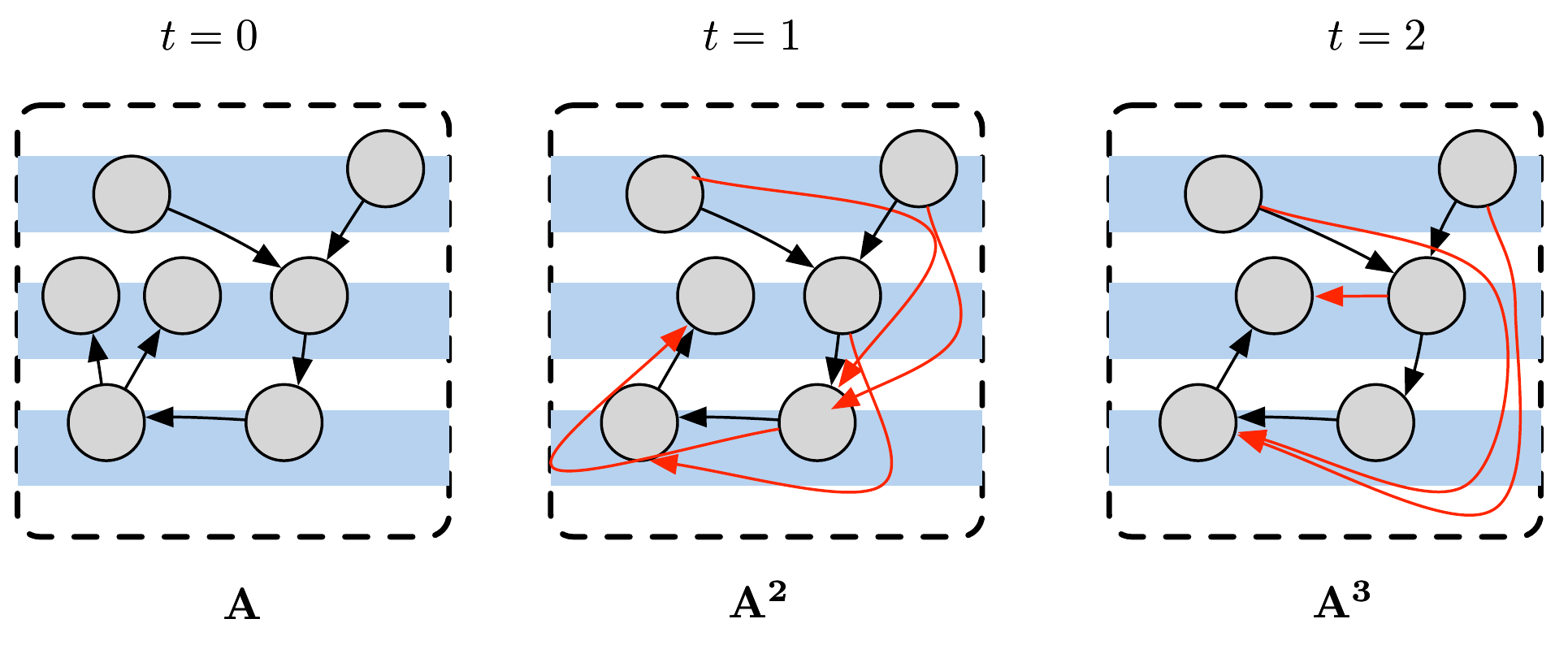}
\caption{\small Propagation in \toolshort\ (shown for a single memory bank across time).  }
\label{fig:prop}
\vspace{-0.1in}
\end{figure}

In this section, we explain propagation with an example. Figure~\ref{fig:prop} shows how propagation happens after every time step. The nodes represent entities corresponding to words in a sentence. As sentences are processed word by word, a directed graph is drawn progressively from $w_{0}...w_{i}...w_{N}$. If sentence $l_k$'s path contains nodes already in the current directed graph, $l_k$ will include said nodes in the its path. After $l_k$ is added to $\textbf{A}$, the model propagates the new update hidden state information $\textbf{a}_i$ among all node states using a GRU. $\textbf{a}_i$ for each node $i$ is equal to the sum of the incoming edges' node hidden states. Additionally, we add a particular emphasis on $l_k$ to simulate recency. At face value, one propagation step of $\textbf{A}$ will only have a reachability of its immediate neighbor, so to reach all nodes, $\textbf{A}$ is raised to a consecutive power $r$ to reach and update each intermediate node. $r$ can be either the longest path in $\textbf{A}$ or a set parameter.

\subsection {Memory bank behavior}
\label{subsec:memorybankbehavior}
\begin{figure}
\centering
\includegraphics[keepaspectratio=true,width=3.2in]{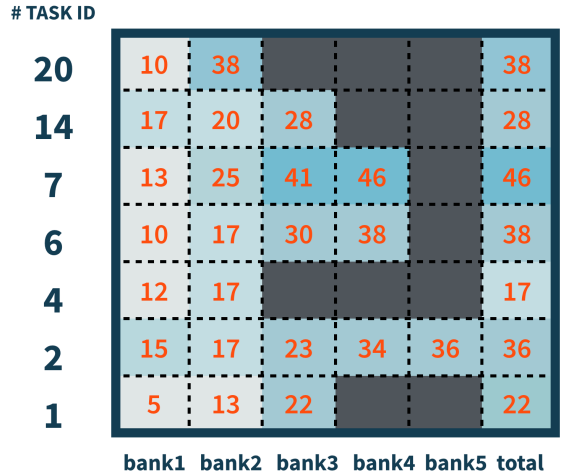}
\caption{\small Heat map showing distribution of entities across various memory banks for simple bAbI tasks. The x-axis shows the task IDs (refer to Table 1 for task details for each ID.)}
\label{fig:heatmap}
\vspace{-0.1in}
\end{figure}

In this section, we understand memory bank behavior of \toolshort. Figure~\ref{fig:heatmap} shows the memory banks and the entity creation for a single story example, for some of the tasks from bAbI. Depending upon the task, and distance from the question \toolshort\ creates variable number of memory banks. The heatmap demonstrates how entities are copied across memory banks. Grey blocks indicate absence of those banks.

\subsection{Algorithm}
\label{subsec:algo}

We describe our overall algorithm in pseudo-code in this section in Algorithm ~\ref{alg1}. We follow the notation as described in the paper.

\begin{algorithm*}
    \caption{AMN($\textbf{S}, \textbf{q}, \textbf{a}$)}
    \label{alg1}
   \begin{multicols}{2}
    \begin{algorithmic}[1]
        \STATE $\textbf{M} \gets \O$
        \FOR{\text{sentence} $s \in$ \textbf{S}}
            \FOR{\text{word} $w \in$ s}
                \STATE $\textbf{D} \gets \text{ENCODE}(w, \textbf{q})$
            \ENDFOR
            \STATE $\textbf{n}_{m_i} \gets \text{SGRU}(\textbf{D})$
            \FOR{\text{memory bank} $m_i \in \textbf{M}$}
                \STATE $m_i \gets \Pi_{au}(m_i, \textbf{D})$
                \STATE $m_i \gets \Pi_{prop}(m_i)$
                \STATE $m_{i+1} \gets \Pi_{move}(m_i, \textbf{n}_{m_i})$
                \STATE $\textbf{n}_{m_i} \gets \text{SGRU}(\textbf{D}, \textbf{n}_{m_i})$
                \IF{$i = |\textbf{M}|$ and $\Pi_{new}(m_i)$}
                    \STATE $\textbf{M},p \gets [\textbf{M}, m_{i+1}]$
                    \STATE Repeat 8 to 11 once
                \ENDIF
            \ENDFOR
      \ENDFOR
         \STATE $\hat{\textbf{a}} \gets \text{DECODE}(\textbf{M}, \textbf{q})$
    \end{algorithmic}
    \end{multicols}
    \end{algorithm*}

    \begin{algorithm}
    \end{algorithm}

\end{document}